\documentclass{article}

\usepackage{microtype}
\usepackage{graphicx}
\usepackage{subfigure}
\usepackage{booktabs} 

\usepackage{hyperref}

\usepackage[accepted]{icml2023}

\usepackage{amsmath}
\usepackage{amssymb}
\usepackage{mathtools}
\usepackage{amsthm}

\usepackage[capitalize,noabbrev]{cleveref}

\theoremstyle{plain}

\theoremstyle{definition}

\theoremstyle{remark}

\usepackage[textsize=tiny]{todonotes}

\icmltitlerunning{Pattern Attention Transformer with Doughnut Kernel}

\begin{document}

\twocolumn[
\icmltitle{Pattern Attention Transformer with Doughnut Kernel}



\icmlsetsymbol{equal}{*}

\begin{icmlauthorlist}
\icmlauthor{WenYuan Sheng}{yyy}
\end{icmlauthorlist}

\icmlaffiliation{yyy}{Undergraduate Student of Information Technology, Beijing University of Tehcnology, Beijing, China}

\icmlcorrespondingauthor{WenYuan Sheng}{wenyuansheng30@163.com}

\icmlkeywords{Machine Learning, ICML, Transformer, Attention, Image Classification}

\vskip 0.3in
]



\printAffiliationsAndNotice{}  

\begin{abstract}
We present in this paper a new architecture, the Pattern Attention Transformer \textit{(PAT)}, that is composed of the new doughnut kernel. Compared with tokens in the NLP field, the canonical Transformer\cite{vaswani2017attention} in computer vision has the problem of handling the high resolution of pixels in images. In ViT \cite{dosovitskiy2020image}, an image is cut into square-shaped patches. As the follow-up of ViT, Swin Transformer \cite{liu2021swin} proposes an additional step of shifting to decrease the existence of fixed boundaries, which also incurs `two connected Swin Transformer blocks' as the minimum unit of the model.
Inheriting the patch/window idea, our doughnut kernel enhances the design of patches further. It replaces the line-cut boundaries with two types of areas: sensor and updating, which is based on the comprehension of self-attention (named QKVA grid). The doughnut kernel also brings a new topic about the shape of kernels beyond square. To verify its performance on image classification, PAT is designed with Transformer blocks of regular octagon shape doughnut kernels. Its architecture is lighter: the minimum pattern attention layer is only one for each stage. Under similar complexity of computation, its performances on ImageNet 1K reach higher throughput \textit{(+10\%)} and surpass Swin Transformer \textit{(+0.8 top1-acc)}.
\end{abstract}

\section{Introduction}
\label{intro}

At the previous time, AlexNet \cite{krizhevsky2012imagenet} and its convolution neural network (CNN) architecture ruled computer vision. The CNN architecture has suitable features that make the models work well on 2D images, like shift, scale, and distortion invariance. While in recent, Transformer \cite{vaswani2017attention} has dominated the field of natural language processing (NLP). There were numbers of directions on how to import Transformer into computer vision \cite{carion2020end,wu2021cvt}.

In the direction of ViT, its vision Transformer is based on the patching action, which changes the high resolution of pixels into token-like units. We notice that the non-overlapping feature between patches decreases models' performance. Starting from this point, we try to design a new kernel, called doughnut. The doughnut kernel is adapted with self-attention in the canonical Transformer. It owns two main differences: abandoning certain trained units in a patched matrix to depart the self-attention into the sensor and updating two processes, and the variety of shapes to the self-attention area. To the first difference, we observe that the process around Q, K, and V matrices makes every unit on the output matrix to be affected with all units in the patch. It provides the basis for our new kernel that in one patch, we use a wider attention area to offer ample local units, but update limited core units. Then it is natural to decide how the sensor area and the updating area arrange in a single patch, which brings our second difference. In our experiments, these two areas are central and symmetrical. Hence, we design the kernel with two cycles of regular octagons. 

To test the validity of the doughnut kernel, we implement it on a testing model, the Pattern Attention Transformer, which highlights that attention patches could composed to different patterns. The regular octagon PAT surpasses the previous state-of-the-art result (Swin Transformer \cite{liu2021swin}) on ImageNet-1K image classification. 

In summary, our proposed doughnut kernel equips benefits from both CNNs and Transformer: local receptive fields and global context fusion. With conceptual generalizing, it reaches a superset definition at the kernel level.

\begin{figure*}[ht]
\vskip 0.15in
\centering
    \includegraphics[width=14cm]{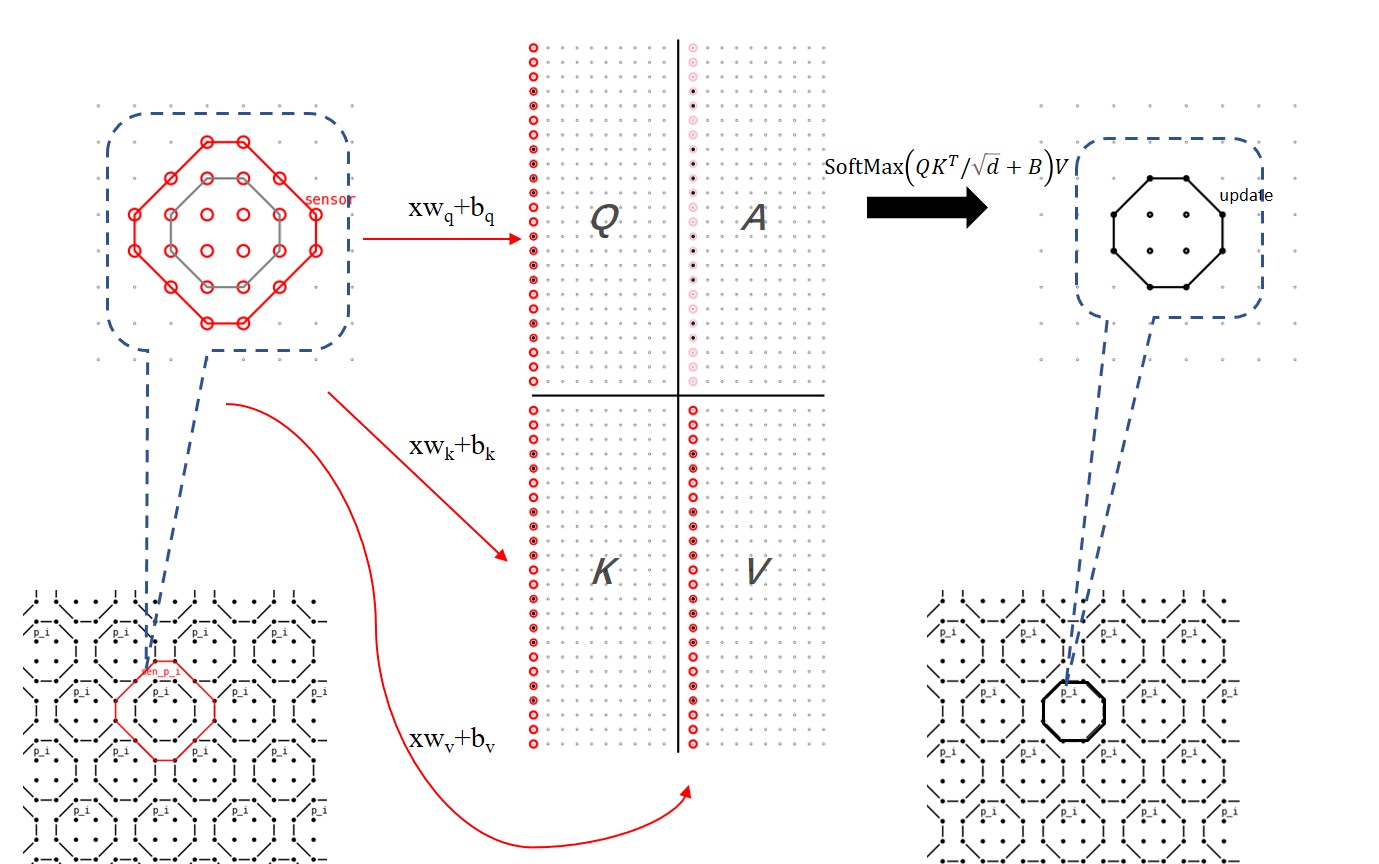}
\caption{To one doughnut kernel, it has 24 units covered sensor area (red) and 12 units covered updating area (black). In the process of self-attention, each line in the sensor area is rearranged in a column, which contributes to the dimension of spatial resolution (here is 24). In canonical square self-attention, the resolution is always $w{\times}h$.}
\label{fig:qkva}
\vskip 0.1in
\end{figure*}

\section{Related Work}
\subsection{Comprehension of self-Attention with QKVA grid}
In canonical Transformer, the self-attention is achieved with three same size matrices, queue(Q), key(K), and value(V). We leave detailed derivation in the Appendix part.
\begin{align}
\label{form:qkva}
A_{(r,c)}=\sum_{j=1}^w \sum_{i=1}^h Q_{(r,j)}K_{(i,j)}V_{(i,c)} 
\end{align}
Here we reach the conclusion that each unit(A(r,c)) on the output (A) of self-attention is traversed with the row(r) in matrix Q, the column(c) in matrix V, and all units in matrix K. In other words, any single attention unit is affected with all units in a patch, which infers the design of the doughnut kernel.

\subsection{Why Doughnut}
In the previous design of ViT, patches are defined to correspond with tokens in the NLP field. Its impressive achievements prove the validity of patching on image classification. Here we propose a new design of patches, the doughnut kernel. In the work of ViT and its follow-ups, the definition of patches solely follows the square shape with manually designed boundaries, which causes two adjacent pixels' positional interactions to be dismissed when they belong to different patches. 

To set overlapping between batches, we use the feature proved by QKVA grid. It notices that some units in patches could take part in the attention calculation but no need to update themselves. Hence in computing self-attention, we depart the process into two steps: sensor and updating, which means each attention block collects and computes from a wider area around it but only reflects at the core area. 
As shown in Figure \ref{fig:qkva}, though all three matrices have $24{\times}feature$ size, we only use the inside 12 units. By dropping other red dots, a wider attention is reflected in a smaller and central area. We also adapt a relative position bias $B\in P^{24\times24}$ \cite{liu2021swin}.

\begin{figure}[h]
\centering
    \includegraphics[width=2.5cm]{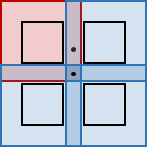}
\caption{The square shape kernel.}
\label{fig:square}
\end{figure}

It is possible to change the shape of doughnut kernels. One of our choices is the regular octagon. Since all units with the same distance from the central point should have similar effects, a more circular patch works better. As shown in Figure \ref{fig:square}, when the updating kernels are not overlapped, the unit at the corner is calculated four times, while the others at the edges are calculated twice by the adjacent kernels. To balance that each unit is averagely covered, we replace the traditional square kernel with circle-like shapes. 
Though the doughnut kernel is inspired by the process of introducing Transformer into computer vision, it equips the features from both the convolutional kernel and the Transformer attention. Compared with CNN’s kernel design, the doughnut kernel reaches a similar overlap between patches but employs self-attention to train a matrix, rather than one point; Also, the updating area of a doughnut kernel should not be overlapped, which corresponds with the certain steps of convolutional kernels.

\subsection{Pattern Attention}
\begin{figure}[h]
\centering
    \includegraphics[width=7.7cm]{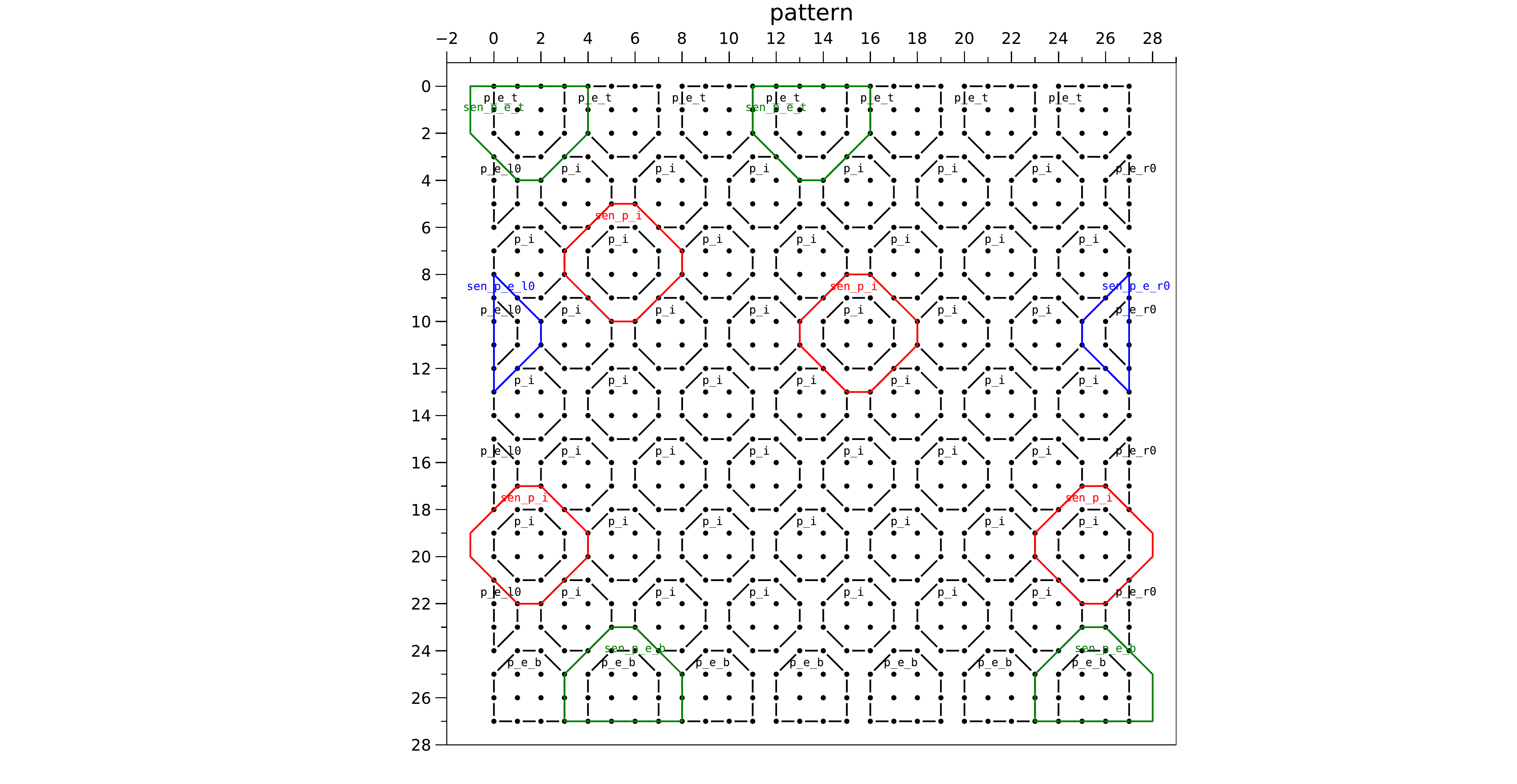}
\caption{The regular octagon pattern. We depict few coloful attention areas.}
\label{fig:pattern28part}
\end{figure}

The name of pattern attention is inspired by the first sight of the analyzed image which is covered with a pattern composed of two types of areas. When we compose various shapes of doughnut kernels, the different patterns leave direct impressions. In our experiments, the regular octagon is better suited for the $224\times224$ size inputs. As shown in Figure \ref{fig:pattern28part}, there are three types of kernels: the regular octagon at inners (p\_i), the trapezoid at horizontal sides, and the irregular pentagon at vertical sides. The additional two shapes compensate for using every unit without zero padding because the regular octagon patches require more than one circle of padding in this size.

\begin{figure*}[ht]
\vskip 0.15in
\centering
    \includegraphics[width=10cm]{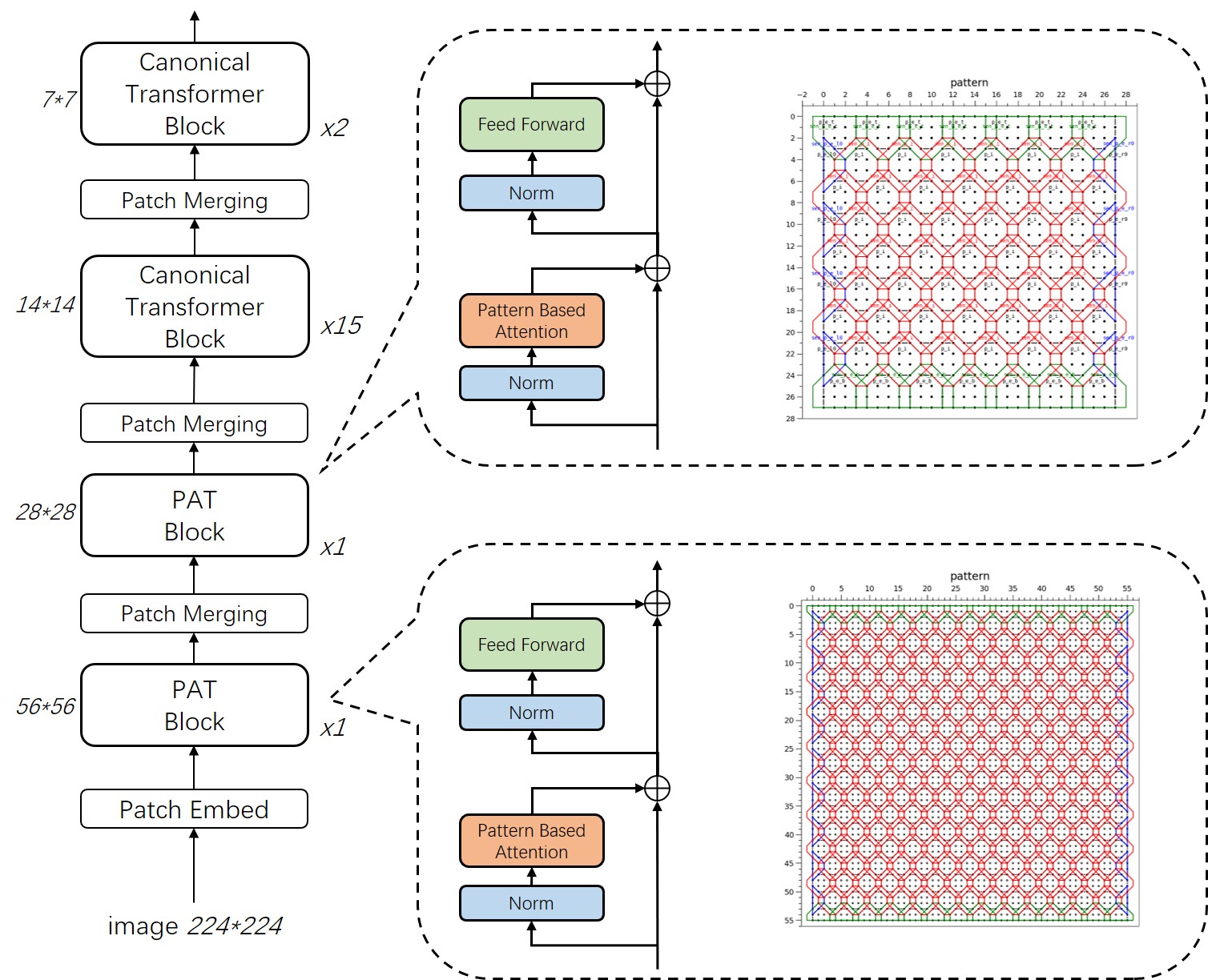}
\caption{The architecture of the regular octagon Pattern Attention Transformer.}
\label{fig:pat1}
\vskip 0.1in
\end{figure*}

\section{Method}
\subsection{Difference of Transformer block}
We keep the four stages architecture from Swin-Transformer, however, all the Transformer blocks are completely replaced. We use the PAT block in the first two stages. As shown in the right side of figure \ref{fig:pat1}, the PAT block has a similar structure compared with the canonical Transformer. The only difference is the pattern-based multi-head self-attention. And in the last two stages, considering the sizes are small enough to do global context fusion, we use the canonical Transformer with its original self-attention. Except for the Transformer blocks, other blocks like patch embed are unchanged from Swin-T.

\subsection{Position Bias}
Generating a position vector (Spatial positional encoding) into the image inputs before Attention processing can be understood as directly adding location information to the image. The topic about this is whether the added location information is fixed or trainable for optimization.

According to our QKVA grid analysis method, we tend to choose $P= Q@K + B$. As shown in Figure 1 and 2, we can divide the transformation process of QKVA into two steps. The first one is $P= Q@K$, the hidden matrix with $pixels\_update \times pixels\_sense$ degrees. Here we notice that the P matrix only contains pixel information but no feature information. By introducing a trainable B-matrix (PixelsUpdate * PixelsSensor), it can be interpreted as the position traversal weight of any Update pixel relative to each pixel in a Sensor. B is trainable, which can be seen as introducing trainable location information for the model. 

In order to reduce experimental expenses, the ablation experiment was not set with high-level configuration settings (96 2-2-18-2). All experiments below are set to 96 1-15-15-1, the conclusions obtained from this level of experiment are equally reliable.
We first set B to be freely adjustable at all points, which we call (Experiment 1) absolute bias (similar description in Swin, unfortunately not presented in the code). We compared the performance of models without and with the Bias matrix(B), and found that with the abs-B, it was 83.06-82.53=0.5 better than without the B. The experimental results confirm the effectiveness of this B-matrix design. With the introduction of abs-B matrix, the parameter number of PAT is larger than Swin Transformer. And to reduce this effect, we try to use shared abs B matrix between multi-heads of one layer. In this condition, the performance reduce 0.1\%.

\begin{figure}[h]
\centering
    \includegraphics[width=2.5cm]{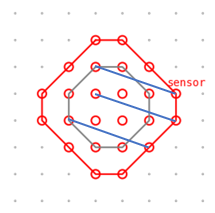}
\caption{The square shape kernel.}
\label{fig:pbias1}
\end{figure}

\begin{figure}[h]
\centering
    \includegraphics[width=2.5cm]{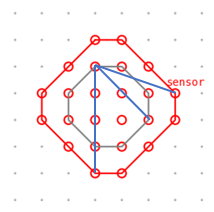}
\caption{The square shape kernel.}
\label{fig:pbias2}
\end{figure}

\begin{figure}[h]
\centering
    \includegraphics[width=2.5cm]{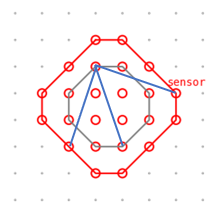}
\caption{The square shape kernel.}
\label{fig:pbias3}
\end{figure}

Through visual observation (Experiment 1) of the data of B in each layer of the absolute bias, as shown in Figure \ref{fig:pbias1}, \ref{fig:pbias2}, and \ref{fig:pbias3}, it provides a very obvious regularity in the model, which further supports the assumption that B reflects a relative position information of pixels. The results of unconstrained training of the model also approaches this relative position relationship. Inspired by the related bias in Swin, we attempt to add constraints to the setting of B matrix, so that the points on it are no longer freely trained. Here we try three constraint schemes between two points, namely, vector, Manhattan Distance, squared Euler distance, as shown in Table \ref{table:bias}.


Furthermore, we noticed in Figure 5 that the regularity is not very strong. Hence we have a hypothesis that the reason for the good performance of Bias with constraints is that for unconstrained Bias, changes in the values of each point during model training will constrain each other, and achieving complete optimization requires more data and training. Adding certain reasonable constraints is equivalent to significantly reducing the number of Bias parameters, which can achieve parameter optimization in less data and training. The parameters in Figure 3 and Figure 4 are $196\times196$ and $49\times49$, while the parameters in Figure 5 are only $24\times12$ (taking the octagon kernel p-i as an example). The much lower parameter number lead to another suggestion that if adding constraints to Bias is certainly the better choice. We also designed further experiments, using abs-Bias for $56\times56$ and $28\times28$ levels, and related Bias for $14\times14$ and $7\times7$ levels The resulting performance is 83.29, which seems to support our hypothesis. We will conduct further experiments based on this setup plan in the future.

\begin{table}[H]
\caption{The vector constraint performs the best, which is the same as the related Bias in Swin Transformer.}
\label{table:bias}
\vskip 0.15in
\begin{center}
\begin{small}
\begin{tabular}{cc}
\toprule
bias constraint & \begin{tabular}[c] {@{}c@{}}ImageNet \\ top-1 acc.\end{tabular} \\
\midrule
$(deltaX, deltaY)$                            & 83.18\%  \\
$\lvert deltaX\rvert + \lvert deltaY\rvert$   & 82.90\%   \\
$deltaX^2+deltaY^2$                           & 83.07\%  \\
\bottomrule
\end{tabular}
\end{small}
\end{center}
\vskip -0.1in
\end{table}
	
\subsection{Block bias}
The above discussion is based on a local image block (a kernel that completes the Attention operation). Training the model to obtain the required location information is beneficial for improving model performance. However, in the same image, local image blocks with the same shape use the same set of parameters, so in terms of positional information, it is not yet possible to distinguish the differences between different local image blocks. Based on this consideration, we added an image block Bias (for a kernel only has one value, or the total parameter number is very small, we use abs) in our experiment.

\begin{align}
\label{form:qkva}
P= Q@K +B\_kernal+B\_block
\end{align}

Through the Ablations experiment observation, the performance has improved by 0.05\%, which is not significant.

\subsection{Winnow computation}
\begin{figure}[h]
\centering
    \includegraphics[width=6cm]{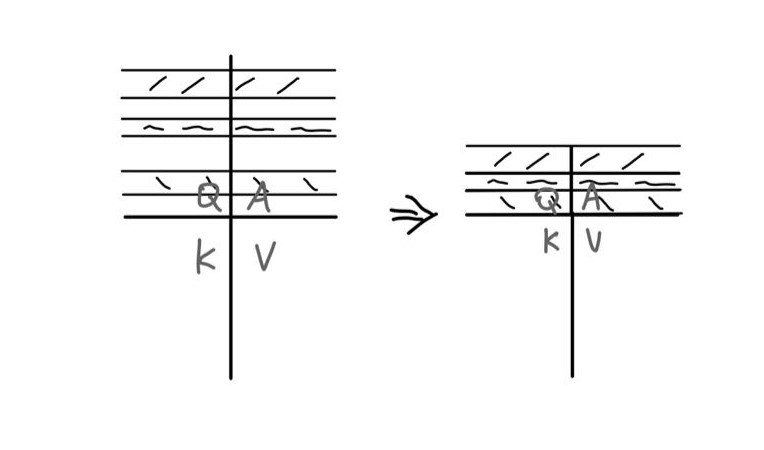}
\caption{The square shape kernel.}
\label{fig:winnow0}
\end{figure}

Figure \ref{fig:winnow0} also provides us with another inspiration. In the process of participating in Attention calculation, only update pixels in Q contribute to update pixels in A (corresponding to only Bias that train corresponding update pixels). If expressed in QKVA-Grid, it can change from Figure 9 to Figure 10, which is called the winnow process. Further, it can lead to winnow attention, which is suitable when update pixels are less than sensor pixels, with less computational complexity and the same computational effect. However, because of the small proportion of computation in $56\times56$ and $28\times28$ layers in PAT, this effect is not obvious. We did not use it in the complete experiment.

\section{Experiments}
Considering the high cost of experiments, we decide to set Swin Transformer as our comparing target, which has reached state-of-the-art performance on ImageNet classification. For this reason, our regular octagon PAT keeps many specific designs of Swin Transformer that are all reflected in our code.
\subsection{Performances of Image Classification}
Swin Transformer implements two types of training on ImageNet 1K: the regular version and the ImageNet 22K pre-trained version. For the cost reason, we select the regular version to compare. Same as Swin, we set AdamW optimizer, GELU activation, 1024 batch, 0.001 initial learning rate with 0.05 weight decay, same numbers of multi-head in self-attention, and any other hyperparameters. To make the comparison fully covered, we also design S and B versions, which correspond same with the model size and computational complexity of Swin-S and Swin-B. In the $224\times224$ size input, our analyzed image sizes in sequence are $56\times56$, $28\times28$, and $14\times14$.

\begin{table}[H]
\caption{Comparisons with Swin Transofrmer. Throughput is measured by setting $AMP\_ENABLE=True$ and $batch\_size=128$ on a RTX3090 GPU. Accuracy of Swin is cited from \cite{liu2021swin}, Table 1.}
\label{table:exp}
\vskip 0.15in
\begin{center}
\begin{small}
\begin{tabular}{cccccc}
\toprule
method & \begin{tabular}[c] {@{}c@{}}basic\\ channels\end{tabular} & \begin{tabular}[c] {@{}c@{}}layer\\ numbers\end{tabular} & \#param. & \begin{tabular}[c] {@{}c@{}}troughput\\ (image/s)\end{tabular} & \begin{tabular}[c] {@{}c@{}}ImageNet \\ top-1 acc.\end{tabular} \\
\midrule
Swin-S & 96 & 2-2-18-2   & 50M & 1016.2  &83.0       \\
Swin-B & 128& 2-2-18-2   & 88M & 728.2  &83.5      \\
\midrule
PAT-S & 96   & 1-1-15-2   & 51M & 1108.2 &83.1      \\
PAT-B & 128  & 1-1-15-2   & 87M & 791.8 &83.6      \\
\bottomrule
\end{tabular}
\end{small}
\end{center}
\vskip -0.1in
\end{table}

\begin{table}[H]
\caption{We also experiment a PAT model with exactly same layer numbers as Swin Transformer. In this comparison, we set 32 batchsize and train on a single RTX3080 GPU. (Because of smaller batchsize, Swin Transformer has lower performance than \cite{liu2021swin} presented.)}
\label{table:exp2}
\vskip 0.15in
\begin{center}
\begin{small}
\begin{tabular}{ccccc}
\toprule
method & \begin{tabular}[c] {@{}c@{}}basic\\ channels\end{tabular} & \begin{tabular}[c] {@{}c@{}}layer\\ numbers\end{tabular} & \#param. & \begin{tabular}[c] {@{}c@{}}ImageNet \\ top-1 acc.\end{tabular} \\
\midrule
Swin & 96 & 2-2-18-2   & 49.6M &82.93       \\
\midrule
PAT & 96   & 2-2-18-2   & 50.5M &83.84     \\
\bottomrule
\end{tabular}
\end{small}
\end{center}
\vskip -0.1in
\end{table}

\subsection{Ablations}
\textbf{Position bias}
To further reduce the parameters of bias matrices, we test on leaving one common bias matrix for every heads in a transformer block. The ablation experiment proves its same effects.
\begin{table}[H]
\caption{Three methods share the 96 basic channels and 1-1-15-1 layer numbers. Multi Bias means there is unique bias matrix for each head. Common Bias means only one bias matrix working for all heads. None Bias means discard of relative position bias.}
\label{table:bias}
\vskip 0.15in
\begin{center}
\begin{small}
\begin{tabular}{cccc}
\toprule
method & \#param. & \begin{tabular}[c] {@{}c@{}}ImageNet \\ max top-1 acc.\end{tabular} \\
\midrule
Multi Bias & 43.9M &83.06      \\
Common Bias & 37.5M &82.95      \\
None Bias & 36.9M &82.53      \\
\bottomrule
\end{tabular}
\end{small}
\end{center}
\vskip -0.1in
\end{table}

\section{Conclusion}
In this paper, we present the doughnut kernel with its various choices of shapes. As the testing architecture of doughnut kernel, Pattern Attention Transformer in regular octagon achieves a higher score on ImageNet 1K. People may have great freedom to select the shape of patterns for specific needs when considering their FLOPs are only correlated with the pixel numbers. The selection of different patterns could stand from the view of aesthetics, culture, philosophy, etc. However, the square(rectangle) pattern is still the most straightforward and important selection. And in this case, the PAT in square plays as the superset of both CNN and Transformer. When the updating area shrinks into one point, rectangle PAT becomes a CNN; When the updating area extends to the sensor area, it becomes a canonical Transformer.

We look forward to specifying the better shape of kernels and more beautiful patterns under different fields.

\bibliography{pat}
\bibliographystyle{icml2023}

\newpage
\appendix
\onecolumn
\section{QKVA Grid for Image Attention}
\label{appendix_qkva}
The transformer \cite{vaswani2017attention} model is becoming a more popular resolution to vision tasks. However, people are used to treating the Transformer as a unit block and do not step further to its inside detail. It is needed to explain why the scaled dot-product attention works well beyond natural language processes. If we set a hidden matrix P to separate the formula as $P=Q\cdot K^T$ and $A=P\cdot V$, the traversing process could be traced as the formulas. 
\begin{figure}[h]
  \centering
  \begin{minipage}[t]{0.65\textwidth}
  \centering
  \includegraphics[width=8cm]{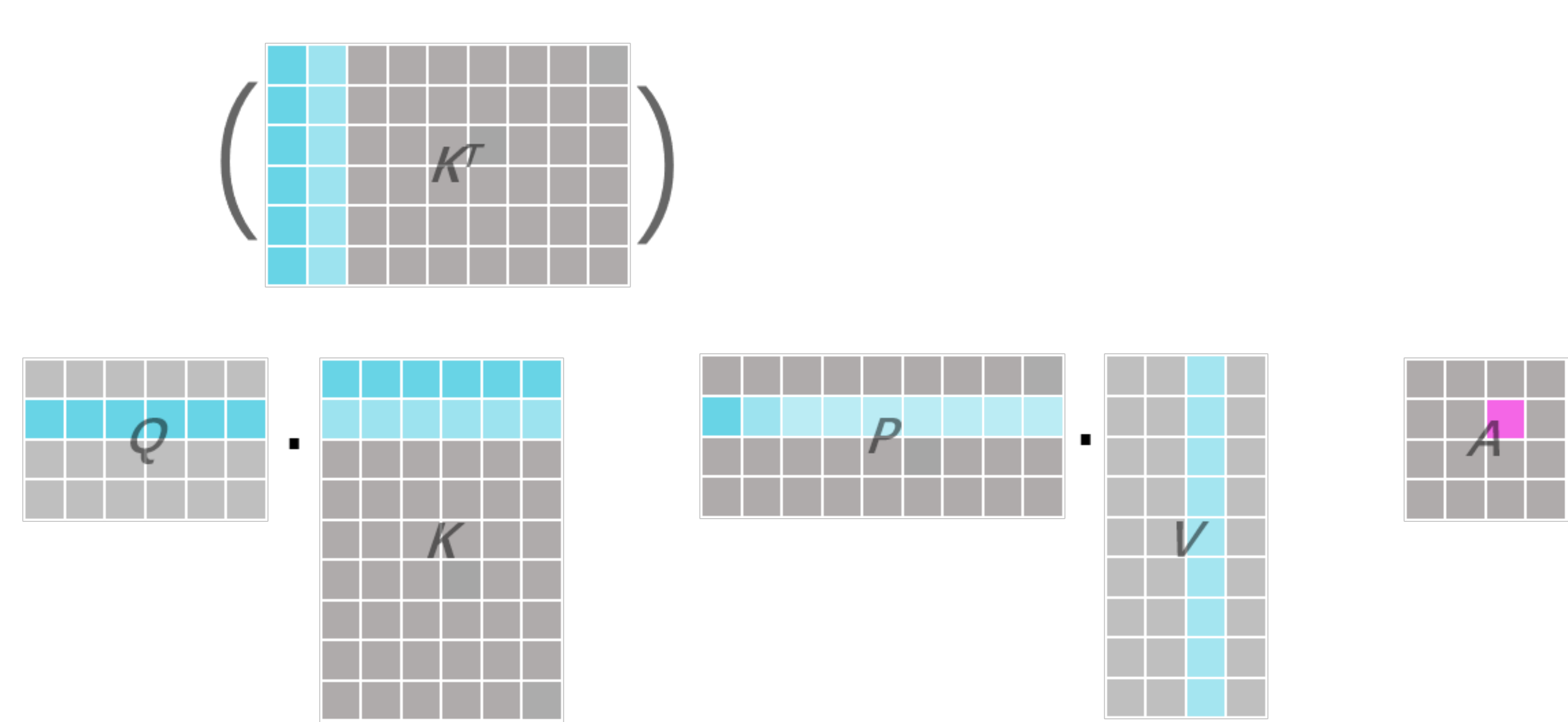}
  \caption{$Q\cdot K^T \cdot V$}
  \label{fig:att1}
  \end{minipage}
  \begin{minipage}[t]{0.3\textwidth} 
  \centering  
  \includegraphics[width=4cm]{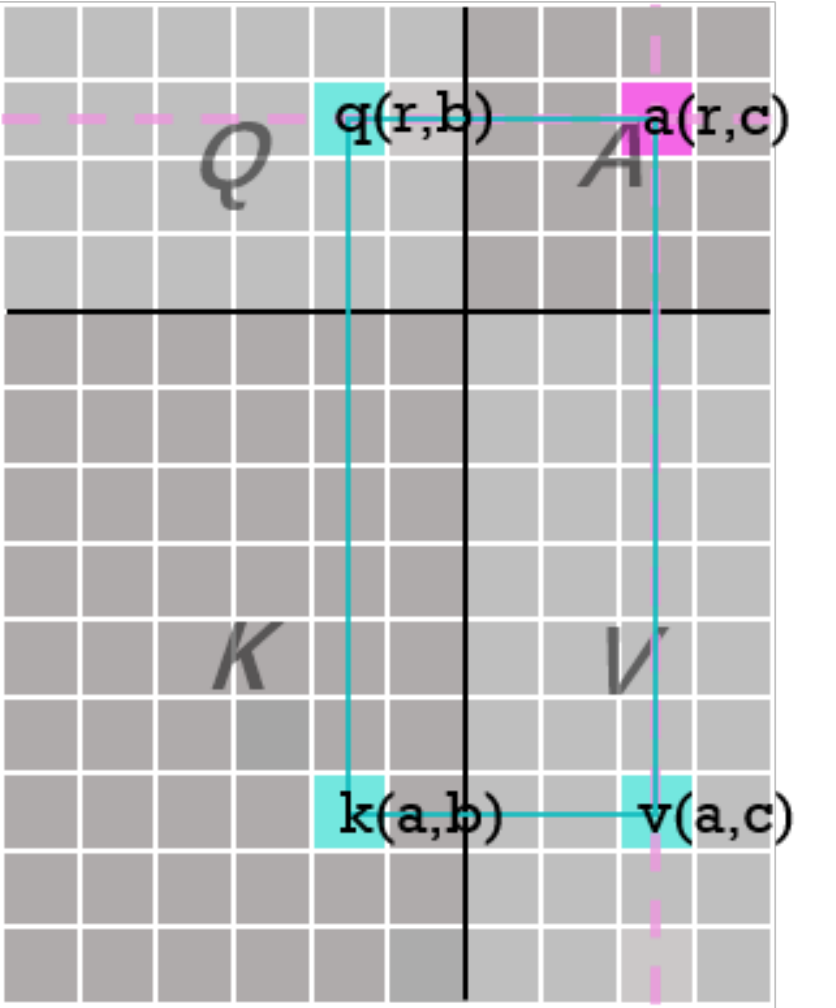}
  \caption{QKVA grid}
  \label{fig:att2}
  \end{minipage}   
\end{figure}
\begin{align}
P_{(a,b)}
=&Q_{(a,1)}\cdot K^T_{(1,b)} + Q_{(a,2)}\cdot K^T_{(2,b)} + \dots + Q_{(a,w)}\cdot K^T_{(w,b)} \\
=&Q_{(a,1)}\cdot K_{(b,1)} + Q_{(a,2)}\cdot K_{(b,2)} + \dots + Q_{(a,w)}\cdot K_{(b,w)} \notag
\end{align}
\begin{align}
Attention_{(x,y)} 
=&\quad   P_{(x,1)}\cdot V_{(1,y)} +              P_{(x,2)}\cdot V_{(2,y)} + \ldots \ldots +   {}           P_{(x,w)}\cdot V_{(w,y)} \\
=&\quad   Q_{(x,1)}\cdot  K_{(1,1)}\cdot V_{(1,y)} + Q_{(x,2)}\cdot K_{(1,2)}\cdot V_{(1,y)} + \ldots + Q_{(x,w)}\cdot K_{(1,w)}\cdot V_{(1,y)} \notag\\
 &+   Q_{(x,1)}\cdot  K_{(2,1)}\cdot V_{(2,y)} + Q_{(x,2)}\cdot K_{(2,2)}\cdot V_{(2,y)} + \ldots + Q_{(x,w)}\cdot K_{(2,w)}\cdot V_{(2,y)} \notag\\
 &+   \vdots         	                                                                                 \notag\\
 &+   Q_{(x,1)}\cdot  K_{(h,1)}\cdot V_{(h,y)} + Q_{(x,2)}\cdot K_{(h,2)}\cdot V_{(h,y)} + \ldots + Q_{(x,w)}\cdot K_{(h,w)}\cdot V_{(h,y)} \notag
\end{align}
\begin{align}
A_{(r,c)}=\sum_{j=1}^w \sum_{i=1}^h Q_{(r,j)}K_{(i,j)}V_{(i,c)} 
\end{align}

If we joint three matrices on the dimensions with the same length, as depicted in Figure \ref{fig:att2}, we will get a general area including attention matrix(A). To any unit(a) on matrix A, its value is contributed by three kinds of units in Q, K, V: with the traversal of units(k) on K, we got the other two units(q\&v) by leading a projection to row-r and column-c. It proves that unit-a collects information from every unit in K. Again, by repetitively traversing all units a, matrix A is acquired. In conclusion, the attention used in Transformer module is repeating the collection of information from orthogonal locations.

\section{Patch/Window Attention}
\begin{figure}[h]
  \centering
  \begin{minipage}[t]{0.3\textwidth}
  \centering
  \includegraphics[width=4cm]{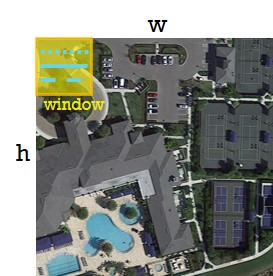}
  \caption{}
  \label{fig:window1}
  \end{minipage}
  \begin{minipage}[t]{0.65\textwidth} 
  \centering  
  \includegraphics[width=8cm]{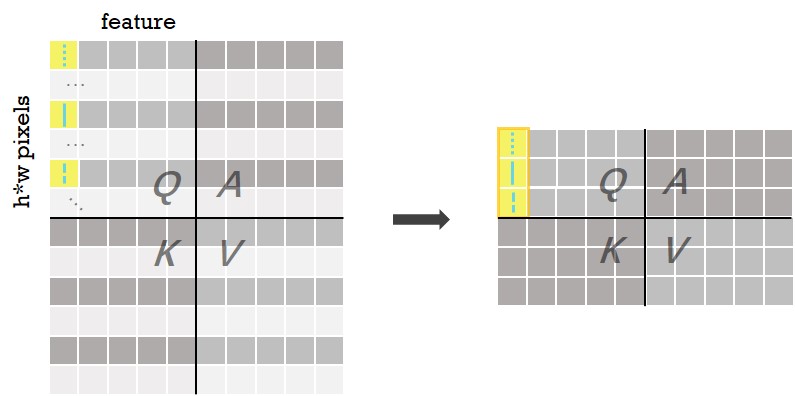}
  \caption{We name the `hw pixels' that is viewed from the $h*w$ image size.}
  \label{fig:window2}
  \end{minipage}   
\end{figure}
In QKVA grid, multi-head groups the matrices in columns(feature dimension), and each column is isolated from the others. Similarly, it is possible to view the QKVA grid in rows, and which derives the idea of updating certain ratio of units in a patch. In figure \ref{fig:window1}, we highlight a yellow patch from a image where each row is named with unique type of blue line. All three rows are reflected in figure \ref{fig:window2}.

What a patch/window works is thinking backward: if we select every certain part of pixels on $w$ dimension(the yellow units in fig \ref{fig:window2}), it would correspond to a square in the image. The square is same as `window' in Swin-Transformer \cite{liu2021swin}. It is the basic unit of localized attention. Based on this view, we may consider the act of abandoning certain rows from QKVA grid, which is the core idea of doughnut kernel.


\end{document}